# Fog Robotics: A Summary, Challenges and Future Scope

Siva Leela Krishna Chand Gudi, Benjamin Johnston, and Mary-Anne Williams

*Abstract*—**Human-robot interaction plays a crucial role to make robots closer to humans. Usually, robots are limited by their own capabilities. Therefore, they utilise Cloud Robotics to enhance their dexterity. Its ability includes the sharing of information such as maps, images and the processing power. This whole process involves distributing data which intend to rise enormously. New issues can arise such as bandwidth, network congestion at backhaul and fronthaul systems resulting in high latency. Thus, it can make an impact on seamless connectivity between the robots, users and the cloud. Also, a robot may not accomplish its goal successfully within a stipulated time. As a consequence, Cloud Robotics cannot be in a position to handle the traffic imposed by robots. On the contrary, impending Fog Robotics can act as a solution by solving major problems of Cloud Robotics. Therefore to check its feasibility, we discuss the need and architectures of Fog Robotics in this paper. To evaluate the architectures, we used a realistic scenario of Fog Robotics by comparing them with Cloud Robotics. Next, latency is chosen as the primary factor for validating the effectiveness of the system. Besides, we utilised real-time latency using Pepper robot, Fog robot server and the Cloud server. Experimental results show that Fog Robotics reduces latency significantly compared to Cloud Robotics. Moreover, advantages, challenges and future scope of the Fog Robotics system is further discussed.**

## I. Introduction

OVER the last few decades, robotics have revolutionised the world by entering into human life [1] and industries [2]. According to BCG statistics in 2014, it was estimated that the global market of robotics could reach $67 billion by 2025 [3]. However, they revised their projection to $87 billion by 2025 in 2017 [4]. So, the importance of robotics is tremendously increasing due to its wider global acceptance [5]. To make these kind of robots smarter, J. Kuffner proposed Cloud Robotics (CR) [6]. This technology enables robots to share their outcomes and to perform faster than usual [7] [8]. As Cloud Robotics is inspired by Cloud Computing, it is facing the problems inherited by Cloud Computing due to the growth of data from internet-connected devices [9] [10]. Also, Cloud Robotics does not have the capability to handle a large number of robots due to its limitations [11]. Their limitations include bandwidth, latency and network congestion at backhaul/fronthaul system [12]. In any case, once the robot is in action to achieve its desired tasks, it needs seamless connectivity. Here, low latency plays a vital role in making robots to finish its mission in a timely manner [13]. However, latency is likely to increase and it can make robots unresponsive [14]. It can also make them do unwanted actions rather than desired which affect the human-robot interaction.

Hence, Fog Robotics (FR) is introduced and coined by S.L.K.C. Gudi et al., to solve the above-mentioned limitations of Cloud Robotics [15] [16]. It is defined as an architecture which consists of storage, networking functions, control with decentralised computing closer to robots [11]. It is comprised of a Fog Robot Server (FRS) and the Cloud while a Sub-Fog Robot Server (SFRS) can be employed if there is a surge in traffic. A basic architecture of FR is as shown in Fig. 1. FRS/SFRS configuration system or the location is adaptable and can be changed based on the demand. Also, FRS shares the robot outcomes with cloud and other robots for achieving the best performance. So, whenever a robot makes an inquiry, FRS solves the issue or pass it to the cloud if it is incapable of handling. In addition, S.L.K.C. Gudi et al., evaluated the FR system by considering an assumption of latency value and analysed the proposed architectures [11]. Later, several researchers proposed Fog Robotics for various applications such as to enhance human-robot interaction [11], object recognition/grasp planning in surface decluttering using deep robot learning [17] and dynamic visual servoing [18]. Also, researchers explored its usage for industrial robotic systems [19]. Besides, few more researchers are working on security issues of Fog Robotics [20] and remote monitoring of robots with coordinated movements using fleet formation techniques [21].

Therefore in this paper, we analyse the practical feasibility of Fog Robotics, challenges and the future scope. Coming to the main contributions, they are as shown below:

- We discuss the necessity of a unique Fog Robotics field and compare it with the applications of Fog Computing.
- We provide a snapshot of three different Fog Robotics architectures and then discuss the importance using Rescue Robots scenario. This discussion includes expectations, problems and solution using the scenario.
- We validate the Fog Robotics architectures using a Pepper robot, AWS server as the cloud from all over the world and a Fog Robot server with real-time latency as a parameter.
- We discuss the advantages, challenges and future scope of Fog Robotics which can open new research directions.

*This research was supported by Innovation and Enterprise Research Laboratory (The Magic Lab), Centre for Artificial Intelligence, University of Technology Sydney and ARC, Australia

S.L.K.C. Gudi, B. Johnston, M.A. Williams are with the Innovation and Enterprise Research Laboratory (The Magic Lab), Centre for Artificial Intelligence, University of Technology Sydney, 81 Broadway, Sydney NSW 2007, Australia. (e-mail: 12733580@student.uts.edu.au, benjamin.johnston@uts.edu.au, mary-anne.williams@stanford.edu).

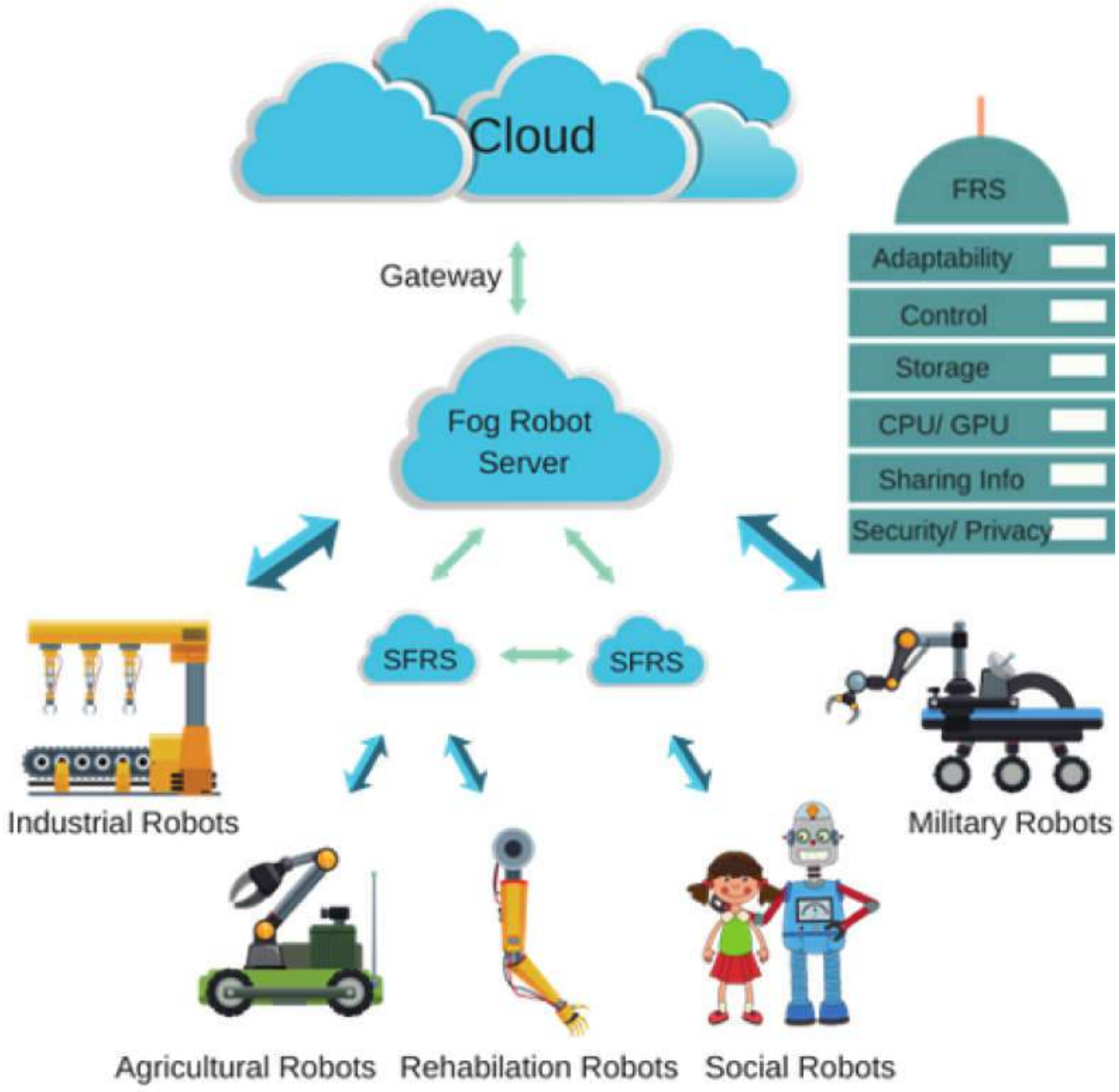

Fig. 1. Architecture of Fog Robotics [11]

Our main goal is to provide a summary of Fog Robotics by demonstrating the importance and checking the feasibility for wide adoption of the Fog Robotics. Coming to the paper, Section II shows the need for a unique Fog Robotics field instead of considering it as Fog Computing based robotics along with a comparison of applications. Next, we describe the architectures and a *Rescue Robots* scenario in Section III. Later in Section IV, we discuss the evaluation setup and results of the architectures. Subsequently, we present the advantages, challenges and future scope of Fog Robotics.

## II. Why Fog Robotics?

The necessity of Fog Robotics instead of Cloud Robotics is clearly demonstrated by Gudi et al., with a comparison in-between them [11]. So, in this section, we discuss on why there is a need for specific field dubbed Fog Robotics instead of considering it as Fog Computing based robotics with respect to applications. Reasons of a robot include but not limited to the working conditions in near real-time environments and the computing power. For instance, if IoT devices are utilising Fog Computing then the equipment considered can be of the low specification whereas for robotics, the requirement should be high due to the massive usage of AI/ML algorithms [22] [23] [24]. However, Fog Robotics shares some of the characteristics of Fog Computing such as the deployment of fog servers and low latency communication. In addition, type of storage, response time, topology, coverage and security protocols are based on the concepts of Fog Computing [25] [26].

TABLE I
Comparison Between Applications of Fog Robotics and Fog Computing

| Parameters | Fog Robotics | Fog Computing |
|---|---|---|
| **Storage** | High | Low |
| **Storage Type** | Transient | Transient |
| **Location** | Distributed | Distributed |
| **Response Time** | Milliseconds | Milliseconds |
| **Topology** | Mostly one hop | Mostly one hop |
| **Coverage** | Local | Local |
| **Security Protocols** | Specific | Specific |
| **CPU/Number of Cores** | High | Low |
| **Number of Tasks** | High | Low |
| **Power Consumption** | High | Low |
| **GPU** | High | Low |
| **Latency/Jitter** | Unacceptable | Acceptable |
| **Mobility** | Unstable | Stable |
| **Real-time Interaction** | Highly Required | Less-likely Required |
| **Bandwidth** | High | Low |
| **Data Transfer Rate** | High | Low |

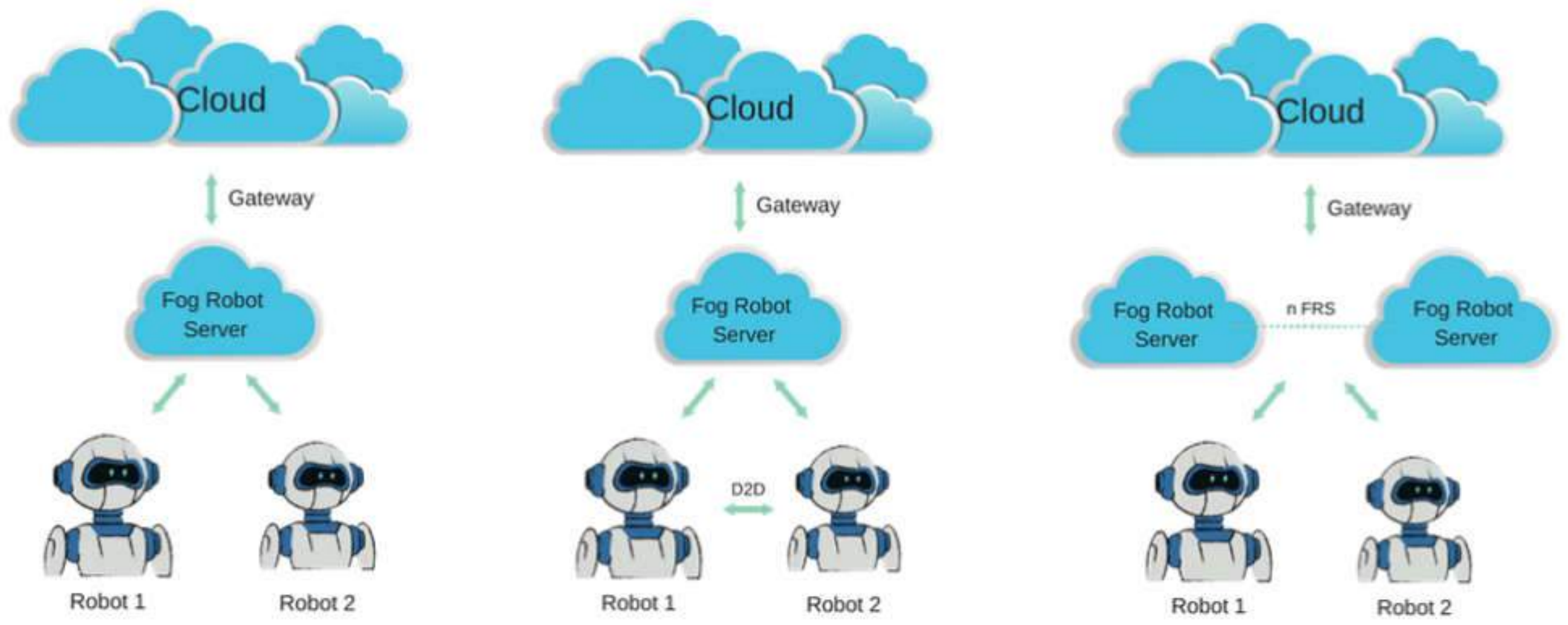

Fig. 2. Case A) Basic FR Architecture, Case B) FR Architecture with D2D Communication, Case C) FR Architecture with Multiple Fog Robot Servers [11]

Another issue that is concerned about Fog Robotics (FR) system is the need for a high amount of storage with more number of CPU/GPU cores due to its data usage. In contrast, general applications of Fog Computing (FC) does not require such high computing requirements. Also, most robots that are currently available in the market comprise of low-level GPU power and it can limit their potential capabilities. Therefore, robots can utilise GPU of the FR system for most of its tasks while it is impossible for FC. For better understanding, we listed the main differences and similarities between Fog Computing and Fog Robotics in Table: 1. Further, the working process of FR consumes a large amount of energy. On the other hand, FC uses low power as it manages mostly the devices which consume less energy [27]. Besides, latency is unacceptable in FR systems as robots need to perform their tasks concurrently and always assumed as hard real-time systems. This requires the need for higher bandwidth because there will be a rapid data transfer between robots and FR system. Conversely, latency is somewhat acceptable when there are less amount of real-time interactions for FC. Therefore, FC can manage with low bandwidth and data transfer rate. Moreover robots are mobile, moving around from one place to another place to finish their assigned tasks. It makes FR system to have handovers in between FRS and robots. Unlike the situation of robot movements, FC applications are most likely to be immobile, staying at one point while performing its task.

Thus, due to specific needs and standards required by Fog Robotics, they are not in a position to use the existing infrastructure of Fog Computing. Also, based upon the above comparison of Fog Robotics and the Fog Computing applications perspective, we believe that a specific field of Fog Robotics is essential rather than Fog Computing based robotics.

## III. Fog Robotics Architectures

Earlier, Fog Robotics architectures (Fig. 2) are proposed by Gudi et al., [11] and in this section, we provide a snapshot of the three different types of models. In each and every architecture, fog robot server (FRS) provides information to the robots and enquires the cloud based on the unavailability of data. For Case A and Case B architecture, only one fog robot server can be used with multiple robots whereas Case C can utilise multiple fog robot servers. Possessing several FRS, robots can receive information from the adjacent FRS when needed. By comparing all of the models, only Case B can have an advantage of edge processing using Device to Device communication (D2D). It can be applicable based on the distance between the robots. If the distance is short then they can use D2D or else an FRS will be utilised.

Based on the necessity and the area of usage such as homes, hotels, airports or parks, any of the three architectures can be employed. If there is a rise in traffic for a particular area then a sub fog robot server can be introduced. To further examine the importance of Fog Robotics, we explore a realistic scenario in the upcoming section.

### Rescue Robots Scenario

Human-robot interaction can be mostly observable during the robot collaboration and communication either with humans or other robot peers [28] [29] [30]. Therefore, we consider a realistic scenario where rescue robots seek to collaborate with others during a fire mishap assisting fire brigades. For instance, if a fire accident happens, there are various problems that firefighters can encounter. They include visibility issues due to smoke, lack of blueprint of the fire-affected zone, stress, health hazards from proximity such as victims being idle due to a panic attack and unconsciousness [31] [32]. Also, firefighters need to search room by room to ensure everyone is evacuated. This time-consuming process may hinder firefighter saving lives efficiently. Therefore, robots can be utilised for assisting firefighters to work more efficiently such as detecting victims beforehand or to map a certain zone [33] [34]. Due to the advanced technology currently available, robots are able to resist fire and detect victims in a smoke-filled environment during evacuation [35] [36] [37]. Also, devices came into the market that can provide the best indoor global positioning system (GPS) by using long-range technology (LoRa) for precise navigation of robots [38]. This led the researchers to propose solutions earlier for fire accidents using humanoids and other types of robots [39] [40] [41]. Generally, the expectations of using such rescue robots during a fire accident are as shown below.

**Expectations**

- Robots are deployed to the disaster area for tracing victims
- Transmits live video streaming to the fire brigade
- Firefighter remotely operates a robot to carry out missions
- Inspects room to room for victims
- Robots communicate with each other and work collaboratively by sharing information of areas covered during missions
- Robots accomplish the mission by saving people

In order to meet the above expectations, rescue robots move around from one place to another to carry out tasks such as extinguishing the fire and to detect victims through dual interaction. To accomplish the goals during rescue missions, they have to perform several tasks. Few important tasks include communicating with firefighter/victims, live streaming the entire event remotely, identifying victims/obstacles and updating the current location in real-time. To achieve these tasks, a robot needs an additional source of computation power for advanced results mainly when AI/ML models are being used. Right now, they work following the principle of cloud robotics. Few researchers also claim that using cloud robotics for these kind of disaster relief missions can help the robot to rescue people as they become increasingly smarter [42] [43] [44]. Robot employing cloud as assistance requires data to be sent back and forth in between them. For instance, if a robot needs to identify an object then it sends data to the cloud which in return, cloud sends back a response with the detailed information of the object. In actual practice, during this transmission/reception process, a robot enters an idle state until it acquires information from the cloud. However, a robot may put itself at risk of fire. Also, it might delay the process of saving victims. Additionally, many tasks might go erroneous against the expectations. Let us consider a few main problems that might arise while using rescue robots. They are as discussed below.

**What happens if?**

- Sensor interrupts while detecting victims
- Streaming fails due to a network congestion
- Firefighter cannot operate efficiently due to latency
- Robot navigation failure due to the issue of mapping update
- Robot cannot communicate with other robot peers
- Robot itself is at risk of fire

One of the main causes for all of the above-said issues is because of the latency or network hitches of cloud robotics. It can also prevent the robot from completing their designated tasks such as detecting victims, updating the map and disconnecting from other robots within a stipulated time. Besides, latency plays a major role as a lag of few seconds can make a robot to risk itself to fire. It also affects robots from collaboration to communication with humans. Besides, robots should be in a position for making decisions in near real-time. Therefore, a lower latency can solve the problem but there is only a chance of rise due to its inflated level of data usage from high-quality cameras (advanced recognition), sensors and other equipment. Eventually, it creates bottlenecks in the network and cloud robotics will become incapable due to its limit for using a number of robots.

To resolve these issues, we can employ the Fog Robotics architectures discussed in the earlier section. Based on the requirement, an architecture of FR can be utilised and the drawbacks can be solved as shown below:

**Fog Robotics can**

- Help robots to receive needed information from other robots by collecting data from sensors.
- Create a better-enhanced network
- Make the applications run with low latency and high-efficiency
- Make robots to create a map together of the building for collaboration
- Make other robots to search victim robot based on past location
- Make a robot to rescue victim robot using a fire extinguisher

Most of the essential information such as maps and danger zones can be stored on the fog robot server. If in case of heavy traffic, sub fog robot server can be employed. Due to the use of fog robot server, latency can be reduced by increasing efficiency when compared to the cloud as it is closer to the robot. It can also allow robots to collaborate with each other faster by communicating information including the number of victims in rooms, joint mapping and sharing of the scanned areas. Even in the absence of an internet connection, a robot can continue its work autonomously. In addition, as robot shares its previous location with other robot peers, they can rescue victim robot by collaborating among themselves. As a result, firefighters can go directly to the exact location of victims and save lives more efficiently. Finally, we believe that cloud robotics possess many limitations related to latency and the quality of service (QoS) as discussed earlier. Therefore, Fog Robotics can become a supplement to cloud robotics or act autonomously for maintaining advanced human-robot interaction.

## IV. Results

Based on our discussion earlier, we believe that latency is causing a major concern for degrading the performance of robots. Hence, we investigate the latency affect on Fog Robotics. Before, preliminary results of Fog Robotics(FR) architectures are discussed by Gudi et al., [11]. They claim that FR performs better than Cloud Robotics (CR) using an assumption of latency value. Therefore in this section, we discuss the advanced analysis of results using real-time latency. For examining the FR architectures, let us consider a social robot Pepper [45], a Fog robot server (FRS) and the Cloud. Firstly, a scenario of FR is considered as shown in Fig. 3. For Cloud, we considered Amazon Web Services (AWS) servers [46] from various locations of the world. They include Australia (Sydney), South Korea (Seoul), Singapore, India (Mumbai), Germany (Frankfurt), United Kingdom (London), South America (Sao Paulo), United States of America (Ohio) and Canada (Central) while a local server is regarded as FRS.

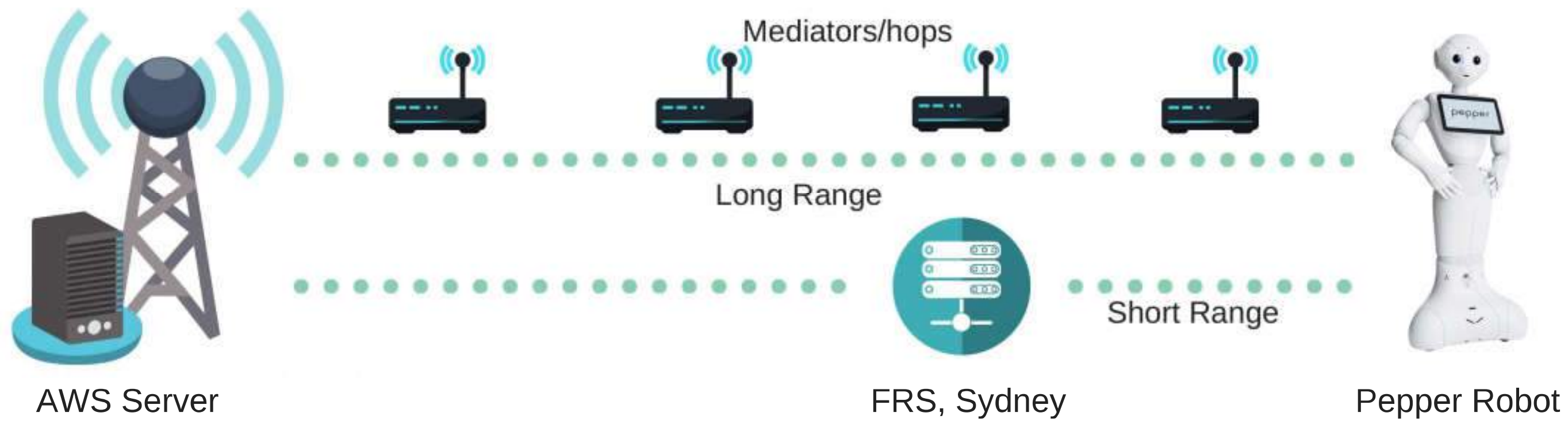

Fig. 3. Architecture of Fog Robotics w.r.t AWS and Pepper Robot

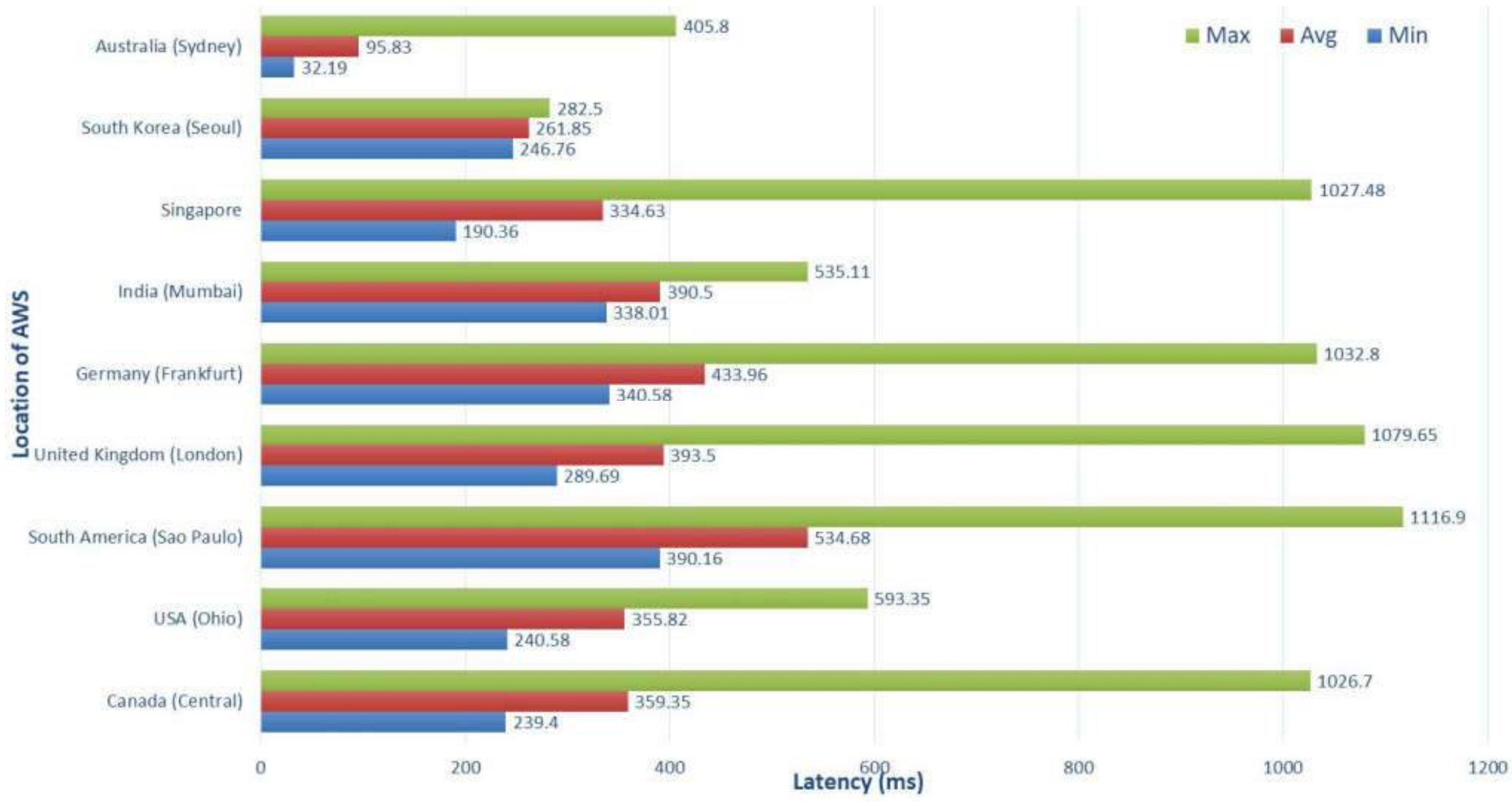

Fig. 4. Latency: AWS w.r.t Robot.

Usually for any kind of scenario, mostly robots exchange information such as images, maps and analysis of speech. This is processed in the form of data either with fog robot server or the cloud. Thus, we considered sending packets of data for calculating the latency. For better understanding, latency is tested with the help of the Pepper robot and are plotted as shown in Fig. 4. Based on the obtained latency results of cloud sever concerning the robot, we can observe different latency values across various countries. For validation purpose, only the highest, average and lowest latency of particular countries are considered. As an average after several attempts, we can see that South America (Sao Paulo) has the highest latency of maximum 1116.9ms, an average of 534.68ms and minimum of 390.16ms. On the other hand, lowest latency is seen at Australia (Sydney) with 405.8ms as maximum, 95.83 as average and 32.19ms of minimum. Alongside, a median of latency is observed at South Korea (Seoul) with 282.5 as maximum, 261.85 as average and 246.76 as a minimum. For performance evaluation of FR, an iFogsim toolkit [47] is chosen for predicting the latency with various conditions. Further, the impact of latency on the proposed three

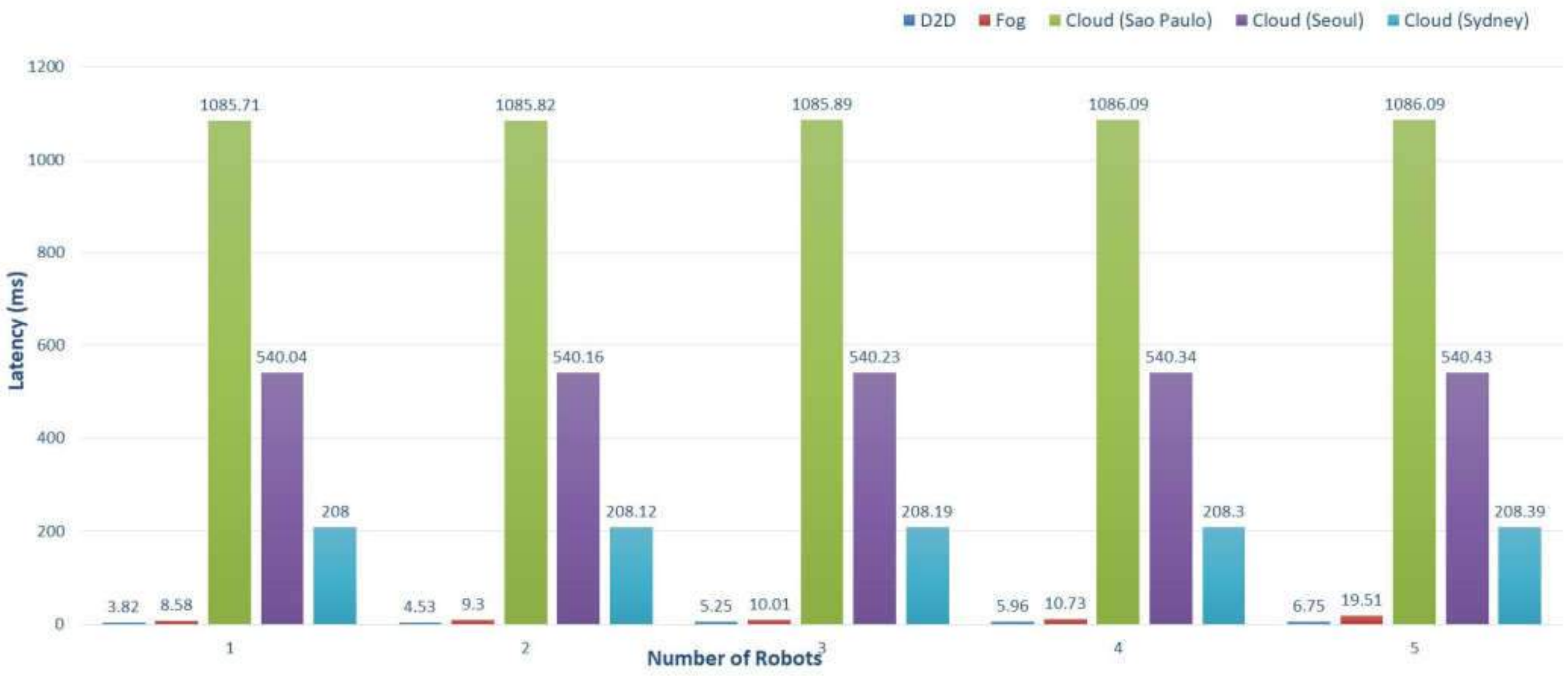

Fig. 5. Results of Architecture A/B Scenarios

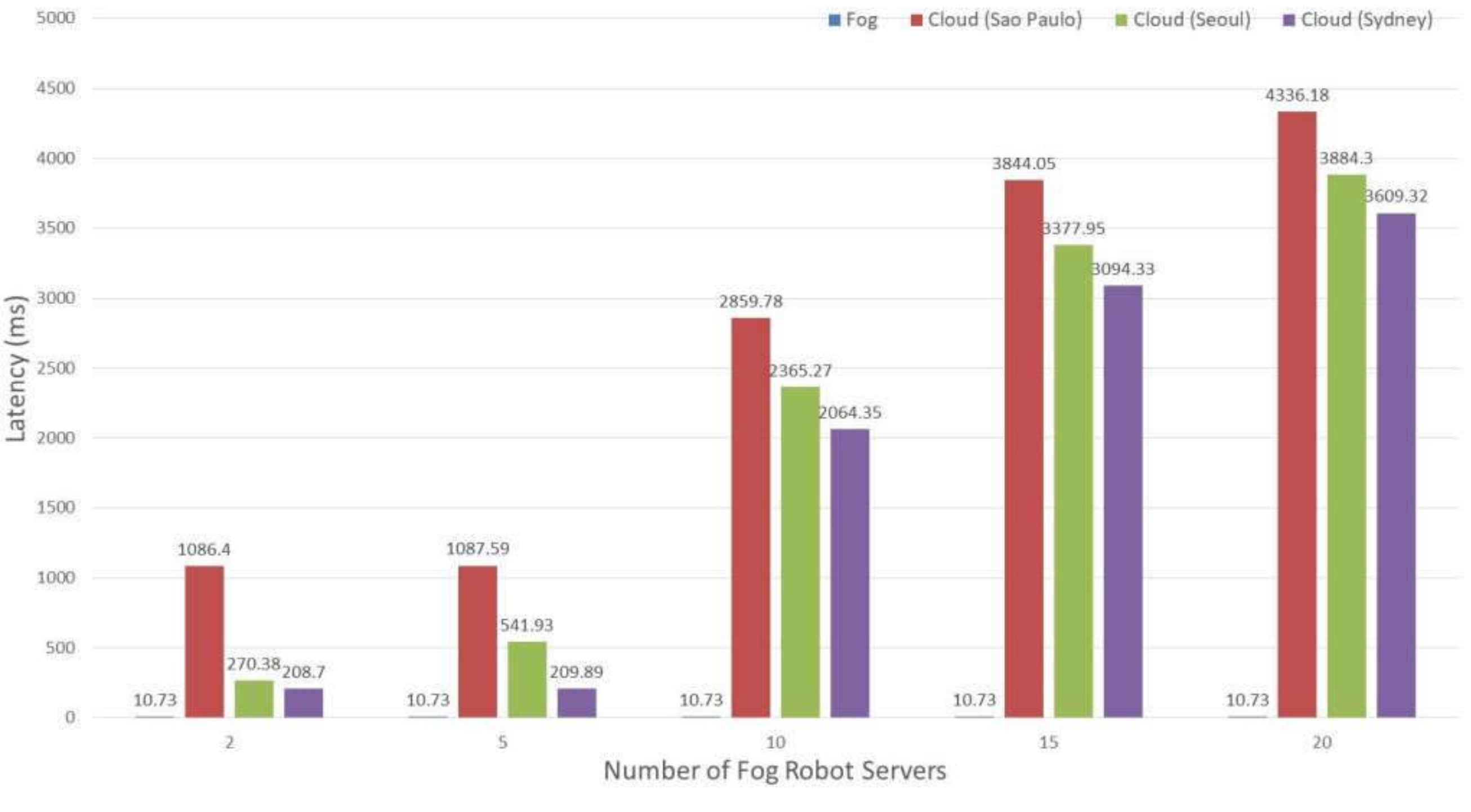

Fig. 6. Results of Architecture C Scenario

architectures are as shown below.

## A. FR Architectures(A/B)

For evaluating the Fog Robotics (FR) architectures (A/B), we chose to validate the latency with a variation of 1-5 robots. These robots send packets of data to Fog Robot Server and the Cloud. Upon measuring the latency w.r.t architecture description, we can say that the latency of FR raised from 8.58ms to 10.73ms hiking to 19.51ms. Also, a sudden spike for FR shows tolerance by a rise in the usage of robots. While D2D has an increase in latency from 3.82ms to 6.75ms with an internal lag at 2ms. Coming to the Cloud, there is a small rise of latency from 1085.71ms to 1086.09ms for Sao Paulo, 540.04ms to 540.43ms for Seoul, 208ms to 208.39ms for Sydney. As the cloud has a capacity for communicating with more than five robots, results have not changed much except to a few milliseconds. While the latency value is more for the cloud when compared to FR as cloud servers are located elsewhere in the world. Finally, a comparison between the architectures (A/B) results is as plotted in Fig 5.

### B. FR Architecture(C)

For architecture C, multiple fog robot servers with a number of robots are connected for advanced communication in between FRS and robots. For experimentation, FRS from two to twenty is chosen consisting each of four robots. For ease of understanding and calculation of results, 2, 5, 10, 15, 20 number of FRS are considered. By examining the obtained results, FR system has retained latency of 10.73ms even when there is a rise in FRS and robots. It is constant because FRS is closer to a robot and have the power of managing robots. CR scenario maintained latency until it can manage but increased when it exceeded its limit. Observed latency values are from 208.07ms to 3609.32ms for Sydney, 270.38ms to 3884.3ms for Seoul and 1086.4ms to 4336.18ms for Sao Paulo.

By analysing the results, we can say that due to the rise of traffic on the cloud at some particular point, latency is increased. Further, they are as shown in Fig. 6. For our evaluation, only particular latencies are considered while in reality, many disruptions can happen in the network. These disruptions can even increase the latency than usual impacting on human-robot interaction. Ultimately, we can say that Fog Robotics perform far better than the Cloud Robotics and plays a major role for the performance of robots.

## V. Advantages

In this section, we demonstrate the main advantages of Fog Robotics (FR) which can impact robots in diverse ways. Accessing data closer to the user and housing essential information, FR consists of many strengths. Some of the benefits are

- Fast response time with low latency
- Avoids bottlenecks and burden on fronthaul and backhaul network system
- Powerful smart enough robots as FR provides required computing, communication and storage capabilities
- Cheaper hardware on robots indirectly offers affordable low-cost robots
- Advanced coordination, collaboration, and communication of robots
- Enhanced battery life by utilising low power
- Increases the durability and life expectancy of robots
- Simple and flexible
- Improves the QoS and QoE
- Reduces the cost of operation

Having the lowest possible latency and avoiding burden in the networks, FR can provide smart affordable robots with advanced collaboration due to its usage of Fog Robot Server (FRS). Also, it can enhance battery life using low power and by sharing the tasks with FRS. Thereby, it increases the life expectancy of robots. Because of its adaptive nature of FRS, the system is flexible with the best QoS/QoE and consequently reduces the operational costs. Even the issues discussed earlier due to Cloud Robotics can be resolved, and an alternative to a cloud system such as additional backpacks can also be avoided [48]. Moreover, proposed models can be applied to different fields of robotics including military, rehabilitation, agricultural, and industrial robots.

## VI. Challenges of Fog Robotics

Along with advantages, there are several unique challenges of Fog Robotics as it is still an emerging new area of research. Some of the main challenges which can open new research directions in the near future are as shown below.

- Maintenance of data
- Availability of FRS/SFRS
- Handover mechanisms in between servers
- Number of current robot users
- Hardware evaluation
- Type of task and estimated processing time
- New/old robot user
- Security protocols
- Status of a robot battery

First, required local data is stored at FRS/SFRS along with cached information. This cached data is collected and should be maintained based on the current local activity of robots. Second, network availability w.r.t FRS/SFRS maintain a vital role as it entails the movement of data/handover mechanisms. Next, the best server must be chosen based on low latency upon the accessibility of servers for best seamless connectivity. This selection will also subject to the over-utilised/under-utilised servers and the number of connected robots. So for monitoring the above-mentioned functions, new algorithms such as time scheduling for advanced integration must be proposed.

Coming to the side of the robot, communicating with servers rely on several conditions. They include the hardware specifications, status of the battery and the estimated processing time for the type of task in reaching its goal. Robots must be registered with the FR system and priority must be given to current users as there might be a chance of malicious robot user. So, to avoid suspicious users, new security protocols must be proposed. Also, the state of battery charge for processing tasks should also be taken into consideration because when there is a low battery, the robot must finish the remaining tasks alone. More rigorous research has to be undertaken for solving the above challenges by proposing novel algorithms. Finally, these can be considered as the future challenges and pave a way to the successful implementation of an effective and robust Fog Robotics system.

## VII. Future Scope of Fog Robotics

In this section, we present the future scope of applying Fog Robotics in different aspects along with possible applications using examples. Discussion includes about the collaboration/communication, privacy, and security.

### A. Collaboration/Communication between Robots

There is a need for robots to share information, communicate and collaborate amongst each other. Let us say two robots named A and B are at two different levels of a hotel. One of the customers on level 8 requested Robot A to deliver a can of coke. Robot A then goes to the fridge, and then realize that coke is not available. It can immediately seek help from robot B on level 9. Upon availability of coke, it informs robot A. Later, Robot A either requests to deliver or goes to his level upon calculating the time needed and distance to accomplish its task. If in case, the customer is going to another level then Robot A transfers information of the customer to Robot B. Information can be a person's name, face, gender, shirt colour, and age. Also, robots must understand the intention of humans by their emotions, predict the actions proactively while working collaboratively [49]. Finally, this can enhance human trust in robots by enhancing the interaction.

Using Fog Robotics, the goal of collaboration/communication can be easily achieved with better responsive rate by acting as a bridge with the existing cloud robotics architecture having a shared control and thus solving the issues of CR. For instance, a robot can save some of the important required information at the fog robot server level and inquire whenever it is in need instead of requesting from the cloud.

### B. Privacy

Privacy is a significant concern in this emerging technological world [50] [51]. A number of breaches are growing at an alarming rate. In future, as the robots increase, the issue of privacy can occur in various situations [52] [53]. If a robot is assisting people along with accepting payments when the deliveries are finished, then there is a need for a robot to gather information about the concerned person or his card details to approve a payment. To perform its task, cameras are must for detecting the people, card/objects, and the environment. If a camera of the robot is hacked, then the robot can be operated as a spy by a hacker to steal information from the people. They can access additional information than required and sneak passwords of their bank during payments.

Even though this situation makes the user frightening, Fog Robotics (FR) solves by taking the sensitive private data to process within FRS instead of requesting access from the cloud. If we consider Cloud Robotics (CR), then in case of a ransomware attack, all of the personal information is automatically leaked to the hackers. But in our scenario, as it is FR, privacy data breaches are hard to occur as they are secured with local new protocols and difficult to be hacked. It can acquire enhanced specific security protocols that are hard to hack than the cloud. Personal/other data can be protected by having ownership. So, Fog Robotics can play a crucial role in safeguarding data when compared to Cloud Robotics.

### C. Security

Robots are prone to be hacked [54] [55]. If a robot with defense equipment is working in a military assisting soldiers, a hacker can turn the robot fight against its own army instead of opponents. No one from the soldiers can be in a position to halt the robot as it is in the hands of a hacker instead of the controller. This might create a massive loss of soldiers life which in turn be a disaster. Robots must be protected from hackers and other security violations.

With the use of Fog Robotics, a new layer of protection can be added in between the cloud and robot. New pins, specific innovative protocols can be added and manageable to the stage of FR. Furthermore, the robot can also keep itself secure. They can communicate by warning between other robots if it sees an accident-prone area or dangerous stuff which could be even water. If in case, a fog robot server is hacked to steal information then only that particular affected Fog Robot server (FRS) can be halted until it is safe from ransomware/malware/virus attack. This makes a small part of the system to disrupt its work while other FRS continues to work on its own task. Finally, Fog Robotics can make the system work without disruptions and can be handled more securely.

## VIII. Conclusion

In this paper, we first discussed the need for specific Fog Robotics field instead of Fog Computing based Robotics along with a comparison. Then, architectures are presented with a realistic rescue robots scenario where they aid the firefighters. Also, the usual expectations of using a robot and problems are mentioned. Later, we showed the importance of Fog Robotics by explaining its usage in the scenario. Next, by considering latency as a parameter, we made a setup using a social robot Pepper, a local server and the amazon web services cloud as a Server. For advanced analysis, latency with respect to robot and cloud server from across the globe is considered. Further, three different architectures are examined by considering the medians of various latency. Finally, we proved that Fog Robotics performance is higher than Cloud Robotics providing the lowest possible latency which can enhance the human-robot interaction. Further, advantages, challenges and future scope of Fog robotics are discussed.

Besides, due to the non-requirement of high-level hardware, robots can become cheaper. It can also reduce the size of robots and process information faster as most of the tasks are sent to FRS/Cloud. In addition, being a supplement to Cloud Robotics, best seamless connectivity can be achieved. Problems discussed earlier such as bandwidth, network congestion at fronthaul and backhaul can be reduced. Similar kind of architectures can be applicable to different areas of robotics. It includes military, telematics, industry and medical robotics. The limitations discussed earlier can tend to open new research directions in the field of Fog Robotics making sure to implement a stable system. In future, we consider exploring other ways of testing the Fog Robotics system with a goal for wide adoption by extending our work in various ways. This includes solving the challenges discussed, testing of new scenarios and implementing various functions to examine the importance of Fog Robotics.

## IX. Acknowledgements

This paper inspects the architectures, need, challenges and future scope of *"Fog Robotics for Efficient, Fluent and Robust Human-Robot Interaction"* written by Gudi, S.L.K.C. et al., and published at IEEE 17th International Symposium on Network Computing and Applications (NCA), 2018 [11].

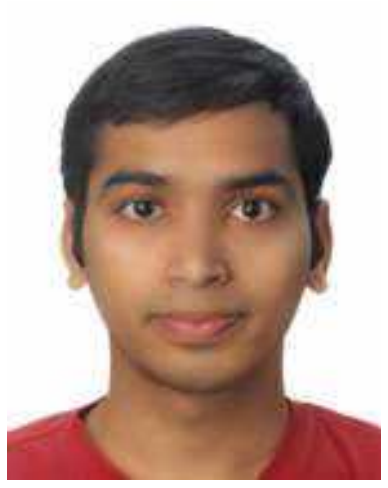

**Siva Leela Krishna Chand Gudi** is a research scholar focusing on Social Robotics at the Magic Lab within Center for Artificial Intelligence of the University of Technology Sydney (UTS), Australia. He is recognized as among the top 200 of the most qualified young researchers globally to attend the prestigious Heidelberg Laureate Forum. He also got invited to Global Solutions Summit which provides policy recommendations to G20/T20 summits and to deliver talks at top-notch venues from MIT CSAIL/Sloan, USA to an audience of Nobel Laureates, Germany. Besides, he worked for various research organizations which include the Commonwealth Bank of Australia (CBA), WENS Lab, Indian Space Research Organization (ISRO), Defence Research and Development Organization (DRDO) along with several patents and publications. Also, he served as a chair/keynote speaker/reviewer for conferences/journals namely RSS 2019, IEEE Access and few more. In addition, he won the world's reputed competitions, various scholarships as well as fellowships including from the Australian Academy of Science.

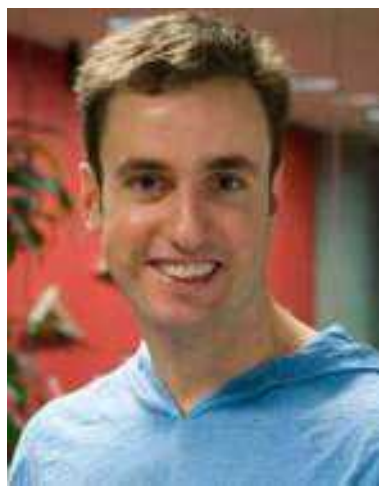

**Benjamin Johnston** is a Senior Lecturer in the Faculty of Engineering and Information Technology at the University of Technology Sydney. He conducts research in commonsense reasoning and social robotics, with a particular interest in using AI to create natural, fluent human-robot interactions. He also teaches classes on social robotics, entrepreneurship and enterprise software development.

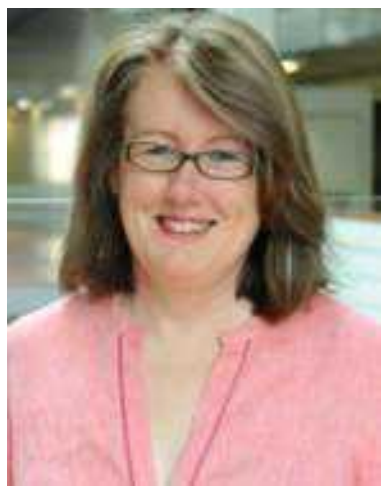

**Mary-Anne Williams** is Director of the Innovation and Enterprise Research Laboratory (The Magic Lab) at University of Technology Sydney. Mary-Anne has a Masters of Laws and a Ph.D. in Knowledge Representation and Reasoning with transdisciplinary strengths in AI, disruptive innovation, design thinking, data analytics, IP law and privacy law. MaryAnne is a Faculty Fellow at Stanford University and a Guest Professor at the University of Science and Technology China where she gives intensive courses on disruptive innovation. Her current research mainly focuses on social robotics while covering wide range of disciplines like belief, perception and risk assessment in robotic agents.